\documentclass{article}

\usepackage{microtype}
\usepackage{graphicx}
\usepackage{subcaption}
\usepackage{booktabs} 
\usepackage{amsfonts}       
\usepackage{nicefrac}       
\usepackage{microtype}      
\usepackage[dvipsnames]{xcolor}         
\usepackage{listings}
\usepackage{fancyvrb}
\usepackage{ml-defs}
\usepackage{cancel}
\usepackage{tocloft}
\usepackage[para,online,flushleft]{threeparttable}

\usepackage{tikz}
\usetikzlibrary{arrows.meta,calc,positioning}
\usepackage{makecell}

\usepackage{hyperref}

\usepackage[accepted]{icml2026}

\usepackage{amsmath}
\usepackage{amssymb}
\usepackage{mathtools}
\usepackage{amsthm}
\usepackage{comment}

\usepackage[capitalize,noabbrev]{cleveref}

\theoremstyle{plain}
\newtheorem{theorem}{Theorem}[section]
\newtheorem{proposition}[theorem]{Proposition}

\theoremstyle{definition}

\theoremstyle{remark}

\usepackage[textsize=tiny]{todonotes}

\icmltitlerunning{Symbol-equivariant
recurrent reasoning models (SE-RRM)}

\definecolor{mydarkgreen}{rgb}{0.0, 0.55, 0.0}

\definecolor{mydarkergreen}{rgb}{0.0, 0.4, 0.0}

\definecolor{mydarkred}{rgb}{0.6, 0.0, 0.0}
\newcommand{\EKComment}[1]{\textcolor{mydarkred}{\textbf{Erich Comment:} #1}}

\definecolor{linkcol}{HTML}{3073AD}  
\definecolor{citecol}{HTML}{3073AD} 
\definecolor{urlcol}{HTML}{3073AD} 

\definecolor{lightblue}{RGB}{32, 194, 217}
\definecolor{jku_red}{RGB}{217, 92, 76}
\definecolor{jku_blue}{RGB}{0, 132, 187}
\definecolor{jku_green}{RGB}{91, 167, 85} 
\definecolor{jku_yellow}{RGB}{241, 188, 63}
\definecolor{jku_cyan}{RGB}{79,176,191}
\definecolor{jku_grey}{RGB}{125,130,140}
\definecolor{jku_lightgreen}{RGB}{191,206,82}
\definecolor{jku_violett}{RGB}{174,97,157}

\newcommand{\TB}[1]{\textcolor{jku_cyan}{Timo: #1}}

\hypersetup{
   colorlinks=true,
   linkcolor=linkcol,
   citecolor=jku_blue,
   urlcolor=urlcol,
}

\renewcommand{\paragraph}[1]{\textbf{#1}\;}

\begin{document}

\twocolumn[
  \icmltitle{Symbol-Equivariant Recurrent Reasoning Models}
  \icmlsetsymbol{equal}{*}

  \begin{icmlauthorlist}
    \icmlauthor{Richard Freinschlag}{ellis}
    \icmlauthor{Timo Bertram}{ellis}
    \icmlauthor{Erich Kobler}{ellis}
    \icmlauthor{Andreas Mayr}{ellis}
    \icmlauthor{Günter Klambauer}{ellis,kfi}
  \end{icmlauthorlist}

  \icmlaffiliation{ellis}{ELLIS Unit Linz, LIT AI Lab \& Institute for Machine Learning, Johannes Kepler University Linz, Austria}
  \icmlaffiliation{kfi}{Clinical Research Institute for Medical AI, Johannes Kepler University Linz, Austria}

  \icmlcorrespondingauthor{Günter Klambauer}{lastname@ml.jku.at}

  \icmlkeywords{Machine Learning, ICML}

  \vskip 0.3in
]



\printAffiliationsAndNotice{}  

\begin{abstract}
Reasoning problems such as Sudoku and ARC-AGI remain challenging for neural networks. 
The structured problem solving architecture family of Recurrent Reasoning Models (RRMs), including Hierarchical Reasoning Model (HRM) and Tiny Recursive Model (TRM),  
offer a compact alternative to large language models, 
but currently handle symbol symmetries only implicitly via costly data augmentation. 
We introduce \emph{Symbol-Equivariant Recurrent Reasoning Models} (SE-RRMs),
which enforce permutation equivariance 
at the architectural level through 
symbol-equivariant layers, 
guaranteeing identical solutions 
under symbol or color permutations. 
SE-RRMs outperform prior RRMs on 9$\times$9 Sudoku and generalize 
from just training on 9$\times$9 to smaller 4$\times$4 \emph{and} larger 16$\times$16 and 25$\times$25 instances, 
to which existing RRMs cannot extrapolate. 
On ARC-AGI-1 and ARC-AGI-2, SE-RRMs 
achieve competitive performance 
with substantially less data augmentation and only 2 million parameters, 
demonstrating that explicitly encoding symmetry improves the robustness and scalability of neural reasoning. Code is available at \url{https://github.com/ml-jku/SE-RRM}.
\end{abstract}

\section{Introduction}

\begin{figure}[th]
    \centering
    \includegraphics[width=\linewidth]{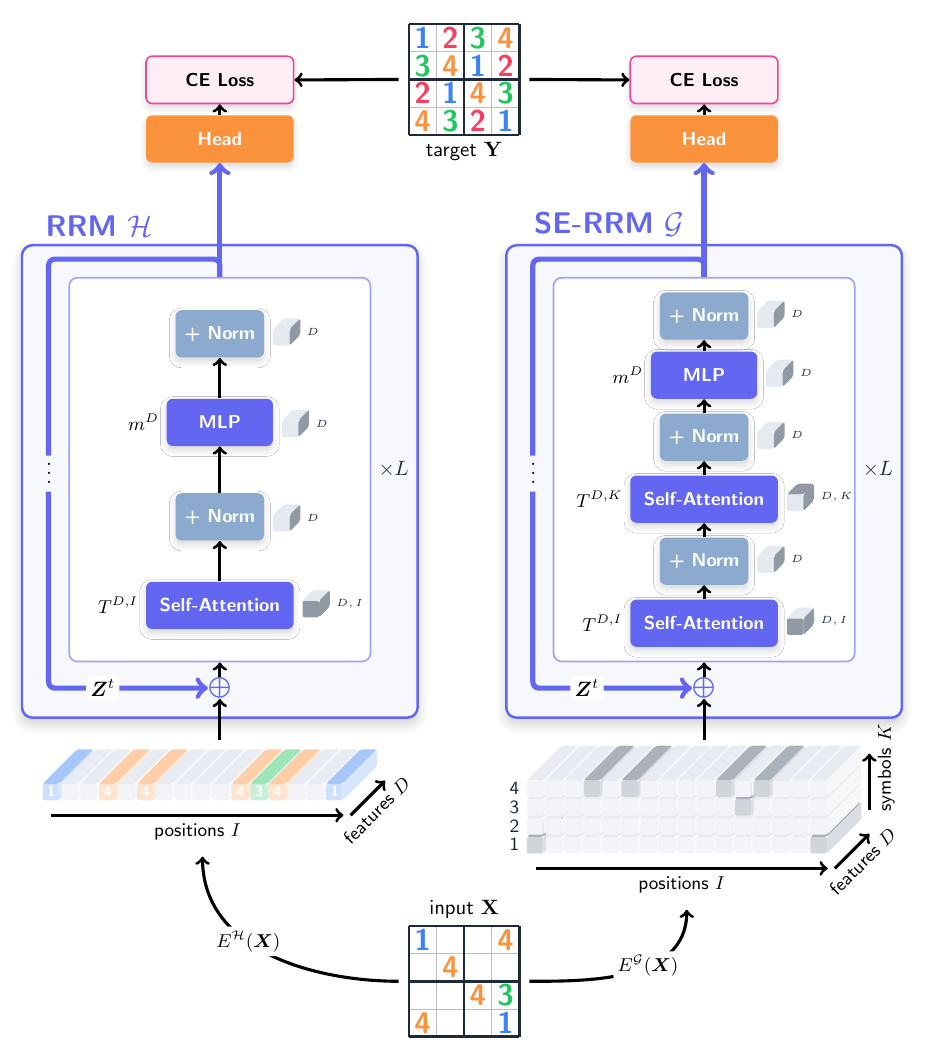}
    \caption{Comparison of a RRM, in particular TRM \cite{jolicoeur2025less}, (left) and the proposed SE-RRM (right) for solving a $4\times 4$ Sudoku. For RRM, the puzzle is encoded with a symbol-specific embedding at every position, resulting in a 2D matrix. In contrast, SE-RRM introduces a third dimension to link positions and symbols and uses the same embedding for all symbols. SE-RRM is symbol-equivariant, since both Self-Attention $T^{D,I}$ and $T^{D,K}$ layers in the transformer block operate subsequently in position and symbol dimension, and the norm, addition, and MLP layers are applied token-wise.}
    \label{fig:main}
\end{figure}

\paragraph{The difficulty of reasoning problems.}
Despite their huge successes, deep neural networks and large language models (LLMs) 
have difficulties at solving reasoning problems \citep{hazra2024can,malek2025frontier, seely2025sudoku}, 
such as undergraduate-level physics \citep{xu2025ugphysics}, 
chemical structures \citep{mirza2025framework,bartmann2026moleculariq}, or, 
real-world clinical scenarios \citep{kim2025limitations}.
Although much of the research community has focused on abstract benchmarks 
such as ARC-AGI tasks \citep{chollet2019measure, chollet2025arc}, 
Sudoku \citep{seely2025sudoku, long2023large}, and maze problems \citep{wang2025hierarchical}, 
these settings can serve as proxies for real-world reasoning challenges, 
including planning, diagnosis, legal interpretation, causal analysis, 
social interaction, and risk assessment. While such benchmarks do not capture the full richness of uncertainty, 
social interaction, or normative ambiguity, 
they isolate the constraint-satisfaction and 
combinatorial structure central to many practical problems \citep{knuth2015}. 
In particular, Sudoku offers a clean testbed for evaluating neural networks
reasoning capacities under hard constraints and structural symmetry.

\paragraph{Symbolic solvers}
Structured problems such as combinatorial optimization and constraint satisfaction 
have traditionally been solved using symbolic methods such as SAT, 
constraint programming (CP), and mixed-integer optimization (MIP) solvers 
\citep{biere2021sat,zhang2023survey,bengio2021machine,kotary2021end}. 
These approaches rely on explicit formulations and combinatorial search, 
and can provide strong correctness and completeness guarantees; backtracking-based Sudoku solvers find a solution when one exists and certify infeasibility otherwise. 
However, symbolic solvers can be computationally expensive: solving generalized 
$n^2 \times n^2$ Sudoku is NP-hard \citep{yato2003complexity} with worst-case complexity 
$O\big((n^2)^{\,n^4}\big)$. 
Learning-based methods can accelerate search by learning solver policies 
\citep{bengio2021machine}, while large language models have also been explored, 
but without task-specific scaffolding their performance on tightly constrained problems 
remains limited \citep{hazra2024can, malek2025frontier, seely2025sudoku}.

\paragraph{LLM-based solutions for reasoning problems.}
Large language models (LLMs) \citep{peters2018deep,devlin2019bert,brown2020language} 
show limited ability to solve reasoning problems expressed in natural language 
\citep{wei2022chain,wei2022emergent,srivastava2023beyond}, 
though this can be improved through 
reinforcement learning and large-scale post-training \citep{guo2025deepseek}. 
Their reasoning is typically evaluated on text-based benchmarks such as BIG-Bench 
\citep{srivastava2023beyond}, which are far simpler than structured combinatorial, 
constraint satisfaction, or SAT problems. 
On harder symbolic tasks, performance plain 
LLM approaches 
degrade sharply: frontier models struggle  
on 3-SAT \citep{hazra2024can} and 
achieve below 5\% on ARC-AGI-2 \citep{chollet2025arc}. 
However, aside from this strategy to use LLMs to directly
predict the output from the input \citep{cole2025don}, LLMs can also be 
used to infer
latent functions, such as Python programs \citep{li2024combining,ellis2020dreamcoder, pourcel2025self,mirchandani2023large}. The latter currently 
appears more promising, since
the strongest ARC-AGI-2 results rely on heavily orchestrated, LLM-guided symbolic pipelines 
with extensive human-designed components and high computational cost 
\citep{poetiq2025arcagi2}. 
As a result, despite their strong performance, 
LLM-based approaches remain 
a limited solution for general reasoning.

\paragraph{Recurrent Reasoning Models (RRMs) for structured problems.}
Recent work has introduced Recurrent Reasoning Model architectures, notably 
Hierarchical Reasoning Model (HRM) \citep{wang2025hierarchical} 
and Tiny Recursive Model (TRM) \citep{jolicoeur2025less}, 
to address discrete reasoning benchmarks such as Sudoku, Maze, and ARC-AGI.
RRMs are a class of neural architectures designed to perform structured, 
multi-step problem solving through iterative refinement. 
At their core, RRMs consist of a neural computation block that is applied repeatedly
in a fixed-point iteration scheme. Rather than producing a solution in a single forward pass, 
the model repeatedly updates its recurrent state, 
allowing outputs to be produced at intermediate steps.
Each block operates on two primary inputs: 
(i) an embedding of the structured problem instance, and 
(ii) a recurrent state that maintains a representation 
of the computation from prior iterations. 
The block produces an updated recurrent state, 
from which task-specific outputs can be derived or decoded at any iteration step.
Training applies a fixed supervision signal, representing the task solution, 
repeatedly throughout the fixed-point iteration. 
After each application of the supervision signal, gradients are detached. 
Then several further iterations of the recurrent block without supervision to refine the recurrent state are performed,
after which the same supervision signal is applied again to the most recent iteration step(s). 
Model parameters are optimized by applying this procedure across a dataset 
of structured problem-solving tasks. 
The iterative updates can be combined with a learned stopping criterion, 
enabling adaptive computation time and dynamic control over the number of reasoning steps per task instance.

\paragraph{Constant-input RNNs.} 
The use of constant inputs in RRMs is not without precedent, 
as constant “plan” vectors were already employed in early 
Jordan-type recurrent networks \citep{jordan1986parallel} to study internally generated temporal dynamics. 
While Deep Equilibrium Models (DEQs) \citep{bai2019deep} similarly rely on fixed external inputs, 
they depart from recurrent formulations by computing equilibria 
through fixed-point solvers and optimizing them via implicit differentiation instead of time-unrolled dynamics. 
As noted in the DEQ literature, this line of work traces back to early equilibrium recurrent networks 
trained via recurrent backpropagation \citep{almeida1987learning,pineda1987generalization,schmidhuber2015deep}. 
Adaptive computation, particularly at test time, serves as a 
natural application setting for DEQs and is likewise a key component of RRMs.

\paragraph{Advantages of RRMs.}
RRMs represent a promising direction for structured reasoning and complement LLM-based methods. As a type of RNN, they can natively perform an arbitrary number of sequential computations, which is required for state tracking and solving many types of reasoning problems \citep{merrill2025illusionofstate}. If LLMs struggle to reason effectively, their robustness to distributional shift may be limited \citep{kim2025limitations}. Given that RRMs are trained with an explicit focus on reasoning tasks, it is plausible that they could offer improved robustness under distributional shift. These characteristics are particularly relevant in domains such as combinatorial optimization and scheduling, where problem sizes and structures vary widely. Although RRMs may introduce higher inference latency due to recurrence, they are often parameter-efficient and can be trained with limited data, without large-scale pretraining or reinforcement learning. Moreover, RRMs decouple reasoning depth from model capacity, enabling compute to scale flexibly at inference time. Some of these advantages have already shown practical promise, for example in insurance pricing with TRMs \citep{padayachy2026tabtrmtinyrecursivemodel}.

\paragraph{Limitations of RRMs.}
A key limitation of existing recurrent reasoning models (RRMs) is the lack of explicit symbol equivariance. 
In many reasoning problems, such as Sudoku, ARC-AGI, and combinatorial optimization, 
symbols are interchangeable and solutions should be equivariant under symbol permutations 
\citep{cohen2016group,bronstein2017geometric}. 
Current RRMs do not encode this symmetry architecturally and instead rely on costly data augmentation, 
which increases sample complexity and can hinder generalization to unseen symbol configurations. 
As a result, RRMs may inefficiently learn redundant distinctions between equivalent symbols, 
limiting both training efficiency and robustness.

\paragraph{Contributions}
    \textbf{a)} We introduce \emph{Symbol-Equivariant Recurrent Reasoning Models (SE-RRMs)}, a new class of recurrent reasoning architectures that are explicitly equivariant under permutations of equivalent symbols, such as digits in Sudoku or colors in ARC-AGI problems. \\
    \textbf{b)} We demonstrate that SE-RRMs reduce reliance on heavy data augmentation. In cases where symbol equivariance is not tackled with extensive data augmentation, SE-RRMs perform substantially better than other RRMs, with only a fraction of trainable parameters. 
    \textbf{c)} We show that symbol equivariance at the architectural level enables the integration of new symbols during inference, while preserving or improving the test-time scaling properties of recurrent reasoning.\\

\paragraph{Related work.} Permutation equivariance has previously been explored in deep sets \citep{deepsets} and graph neural networks \citep{scarselli2008graph, kipf2016semi, defferrard2016convolutional, gilmer2017neural} that are typically invariant or equivariant with respect to permutations of nodes of a graph. Moreover, permutation equivariance with respect to positions is a key property of attention operations as introduced by the transformer architecture \citep{vaswani2017attention}. 
The transformer architecture has already been applied to problems with multiple dimensions. For example, Axial-Attention \citep{ho2019axial} applies one transformer block along the rows and another transformer block along the columns of an image, while Criss-Cross Attention \citep{huang2019ccnet} applies the same principle in the domain of image segmentation. The MSA-Transformer \citep{rao2021msa} employs an attention operation in sequence and another attention operation in residue direction for multiple-sequence alignments of different protein sequences. 
SE-RRM realizes a permutation equivariant-recurrent layer via axial-attention, specifically it applies a self-attention layer across different symbols of the problem.

\section{Symbol-equivariant Recurrent Reasoning Models (SE-RRMs)}

\subsection{Preliminaries: Vanilla RRMs}

\paragraph{Goal.} The aim is to learn a machine learning model for structured problem-solving tasks, such as Sudoku, Maze, or ARC-AGI. We assume access to a training dataset containing representations of tasks together with their corresponding solutions, as well as a separate evaluation dataset. The datasets may consist of tasks that all follow the same set of rules, as in Sudoku, or of tasks governed by different underlying rules, as in ARC-AGI, where multiple task types are present. Depending on the complexity of a task type, multiple task instances are typically required to infer its governing rules. 
For datasets involving multiple task types (e.g. ARC-AGI), typically a task-type identifier is used to distinguish between different types. 

Each task $\BX$ is represented as a tuple of symbols $x_i \in C$, i.e., $\BX = (x_1, \ldots, x_I)$.
The task is assumed to be defined over an underlying structured domain, which is flattened into a fixed ordering to obtain the tuple representation. Each symbol $x_i$ denotes the value associated with the $i$-th element of this flattened domain. The set $C$ consists of $K$ symbol types ($K=|C|$) of the problem-solving task (e.g., the digits of a Sudoku together with a mask symbol) and is often also referred to as the set of colors. The solution $\BY= (y_1, \ldots, y_I)$ of a task is represented analogously as a tuple of symbols $y_i \in C$, sharing the same indexing and flattened domain as $\BX$. Depending on the specific rules of a task, some symbols in $C$ may be absent from the task description, the solution, or both (e.g., the mask symbol in Sudoku appears only in the input). In a dataset of tasks, a task may additionally be enriched with a task-type identifier $p \in \mathbb{N}$. In that way, there is an indication, which tasks follow the same rules and the set of training tasks, then consists of pairs $(\BX, p)$ instead of plain tasks $\BX$.

\paragraph{RRM blocks.} We abstract the principal neural computation block of RRMs, as described above, as a mapping $\mathcal H: \reals^{D \times I} \times \reals^{D \times I} \mapsto \reals^{D \times I}$. We assume the recurrent state at time step $t$ is denoted by  $\BZ^t \in \reals^{D \times I}$. An embedding of the structured problem-solving task is provided by $E^{\mathcal H}(\BX)$. The RRM block then computes the subsequent recurrent state \(\BZ^{t+1}\) from the task embedding $E^{\mathcal H}(\BX)$ and the previous recurrent state $\BZ^{t}$:
\begin{align} 
\BZ^{t+1} = \mathcal H(E^{\mathcal H}(\BX), \BZ^t), \label{RRMfp}
\end{align}

\paragraph{Task Embedding.} An embedding  $E^{\mathcal C}: C^I \mapsto \reals^{D \times I}$, which maps a symbolic task description into a real vector space, is required to process a task.
To construct such an embedding, approaches such as HRM or TRM employ a mapping $C \mapsto \reals^D$ that assigns each symbol within a task alphabet to a \(D\)-dimensional real vector.
This symbol-level mapping is applied uniformly across all positions in the task description (element-wise in the tuple).
The resulting matrix $\BE=E^{\mathcal C}(\BX)$ can be interpreted, in the context of attention-based processing, as a sequence of tokens $\Be_i \in \reals^D$, i.e., $\BE=(\Be_1,\ldots,\Be_I)$.
Since the embedding $\BE$ represents a sequence of symbols originating from a structured domain, it is often beneficial to incorporate positional information. Positional embeddings (e.g., RoPE, \citet{su2024roformer}), can be added to the tokens $\Be_i$ to enable attention-based operations to reason about the relative positions and relationships of symbols within the domain.
Formally, let $E^{\mathcal P}: C^I \to \reals^{D \times I}$ denote a positional embedding mapping that assigns a $D$-dimensional vector to each position $i = 1, \dots, I$ in the original tuple $\mathbf{X}$. The combined task embedding $E^{\mathcal H}$ can then be defined as
\begin{align}
&E^{\mathcal H} :  C^I \mapsto \reals^{D \times I},\\
&E^{\mathcal H} := E^{\mathcal C} + E^{\mathcal P}\nonumber,
\end{align}
where the symbol-level embedding $E^{\mathcal C}(\mathbf{X})$ of a tuple $\BX$ captures the content of the symbols in $\BX$ and the positional embedding $E^{\mathcal P}(\mathbf{X})$ encodes the respective positions.

When a dataset comprises multiple different task types, the embedding function $E^{\mathcal H}$ is extended to encode task-type information and therefore becomes a function of two arguments, namely the task instance $\BX$ and its associated task-type $p$. In principle, this embedding can be realized in different ways. In HRM and TRM, the task type is represented by additional positions ${P}$ whose embedding corresponds to a learned mapping $ \mathbb{N} \to \mathbb{R}^{D \times P}$, which embeds the task type $p$ as a matrix $\in$ $\mathbb{R}^{D \times P}$. The overall task embedding is then obtained by augmenting the original task representation with this task-type embedding, resulting in
\begin{align}
E^{\mathcal H} : C^I \times \mathbb{N} \rightarrow \mathbb{R}^{D \times (I+P)} .
\end{align}
\paragraph{Solution Prediction.}
The overall goal is to predict the correct solution of a task, i.e., to predict the correct tuple of symbols $y_i \in C $ representing the solution. Since this defines a discrete correct/incorrect loss, RRMs employ a differentiable surrogate loss for each position in the solution tuple.
RRMs typically output predicted probabilities for each symbol at every position in the tuple. A categorical cross-entropy loss can then be applied independently to each position.

To obtain the predicted probabilities for the symbols, RRMs typically map the final recurrent state $\BZ^T \in \reals^{D \times I}$ at iteration step $T$ to a logit prediction $\hat \BY \in \reals^{K \times I}$.
A softmax activation function, or a similar function such as stablemax, is applied to these logits, allowing the categorical cross-entropy loss with respect to the true training solution symbols to be computed. For the output mapping $\BO: \reals^{D \times I} \mapsto \reals^{K \times I}$ from the final recurrent state to the logits, a single function $\reals^{D} \to \reals^{K}$ may be applied independently along the position dimension of the recurrent state (i.e., shared across all positions). Usually, this is realized by just using a simple linear mapping (parameterized by a matrix $\in \reals^{D \times K}$).

\paragraph{Overall RRM Architecture \& Deep Supervision.}
The core of the RRM architecture is the application of the subsequent fixed-point operation described in \cref{RRMfp}, where $\BZ^0$ is fixed but randomly drawn and is constant with respect to the position index. Variations of this principle exist: both HRM and TRM can be seen as adaptations of RRMs, as they structure multiple consecutive iteration steps of the block to superblocks. 
Consecutive blocks often omit the task embedding and add recent previous recurrent states to the embedding input, which allows some fixed-point iteration steps to incorporate both the immediately preceding and a more recent earlier state.

An important technique employed in both HRM and TRM is deep supervision. Instead of backpropagating the error from the final prediction through the entire fixed-point iteration scheme, prediction outputs are computed from selected intermediate recurrent states, i.e., $\BO(\BZ^t)$, and the model parameters are updated immediately based on the gradient of the corresponding loss at time $t$: $L(\BY^t, \BO(\BZ^t))$. Gradients are computed only with respect to the last iteration step(s) immediately preceding the recurrent state, from which the output is computed.

\paragraph{Neural Architecture of RRM blocks.}
An RRM block $\mathcal{H}$ realizing one step of the fixed-point iteration, consists of $L$ stacked Transformer-style layers that process the task embedding $E^{\mathcal H}(\BX)$ together with the recurrent state $\BZ^t$, as shown in \cref{RRMfp}. At certain time steps, variants of the architecture may additionally incorporate a recent previous recurrent state $\BZ^{t-r}$ or omit the task embedding, which can be expressed as
\begin{equation}
\BZ^{t+1} = \mathcal H\Big(\mathbf{1}_{\text{emb}}\, E^{\mathcal H}(\BX) + \mathbf{1}_{\text{prev}}\, \BZ^{t-r}, \, \BZ^t\Big), \label{myfprrmb}
\end{equation}
where $\mathbf{1}_{\text{emb}}$ and $\mathbf{1}_{\text{prev}}$ are indicators equal to 1 when the embedding or previous state are included, and 0 when they are omitted. The exact wiring of plain $\mathcal H$ blocks to superblocks in variants of RRM does not appear to be critical, as some works \citep{arc2025hidden, liao2026srm} suggest that it may be replaceable by a stack of transformer layers (with shortcut access to previous layers).

The first layer of an RRM block, which we refer to as the initial layer $\BH_0$, is defined as the sum of all included inputs, i.e.,
\begin{align} 
\BH_0:=\mathbf{1}_{\text{embed}}\, E^{\mathcal H}(\BX) + \mathbf{1}_{\text{prev}}\, \BZ^{t-r} + \BZ^t. \label{eq:rrmb0}
\end{align} 

This initial layer is then processed using the following update scheme
\begin{align} 
    \BH'_{l} &= \text{Norm}\!\left(
        \BH_{l} + T^{D,I}(\BH_{l})
    \right), & \underbar{D} \times \underbar{I} \label{eq:rrmb1} \\
    \BH_{l+1} &= \text{Norm}\!\left(
        \BH'_{l} + m^{D}(\BH'_{l})
    \right) & \underbar{D} \times I  \label{eq:rrmb2}.
\end{align}
for $L$ additional steps, producing the block output $\BZ^{t+1} = \BH_L$. Each RRM block primarily consists of two types of operations. The first are self-attention layers \citep{vaswani2017attention}, denoted as $T^{D,I}: \mathbb{R}^{D \times I} \to \mathbb{R}^{D \times I}$, which operate along the position dimension of the $D \times I$ matrices. The second are position-wise MLPs, specifically SwiGLU modules \citep{shazeer2020glu}, denoted $m^D: \mathbb{R}^D \to \mathbb{R}^D$, which are applied independently to each of the $I$ positions. The utilized $\text{Norm}$ operation is a root mean square normalization \citep{zhang2019root} across the $D$ features, which is also applied independently to all $I$ positions.

\subsection{Notable Properties of RRMs}

The structural design of RRMs endows them with certain properties that can be formally characterized and leveraged in extensions. Among these properties, we note an interesting relationship to positional equivariance: without positional embeddings, RRMs would be equivariant to any permutation applied to the input positions. The following proposition makes this more precise.

\begin{proposition} \label{RRMposeq}
    An RRM block $\mathcal H$ without positional embeddings, as defined in~\cref{eq:rrmb0,eq:rrmb1,eq:rrmb2} is equivariant under permutations $\pi: [I] \mapsto [I]$ of the input positions $\in [I]:=\{1,\ldots,I\}$, i.e.,
\begin{align*}
\Pi_2^\pi\Big(\mathcal H\Big(\mathbf{1}_{\text{emb}}\, E^{\mathcal H}(\BX) + \mathbf{1}_{\text{prev}}\, \BZ^{t-r}, \, \BZ^t\Big)\Big) = \nonumber \\
=\mathcal H\Big(\mathbf{1}_{\text{emb}}\, E^{\mathcal H}(\Pi_2^\pi(\BX)) + \mathbf{1}_{\text{prev}}\, \Pi_2^\pi(\BZ^{t-r}), \, \Pi_2^\pi(\BZ^t)\Big),
\end{align*}
where 
\begin{align*}
&\Pi_a^\pi : \mathcal{S}^{d_1 \times \dots \times d_N} \to \mathcal{S}^{d_1 \times \dots \times d_N}\\
&\big(\Pi_a^\pi(A)\big)_{i_1, \dots, i_N} := 
A_{i_1, \dots, i_{a-1}, \pi(i_a), i_{a+1}, \dots, i_N}.
\end{align*}

\end{proposition}

The proposition is straightforward to show. It relies on the equivariance property of self-attention \citep{ilse2018attention,lee2019set}. Further, since MLPs and the $\text{Norm}$ operation are applied independently to each position, they naturally fulfill equivariance wrt. position permutations. A similar argument holds for the residual additions. Since the pure symbol embedding $E^{\mathcal C}$ is also applied independently to each position, it can be shown that a whole RRM block is position-equivariant by induction. We can even further extend this equivariance property to a whole RRM architecture without the use of any positional embeddings, considering that $\BZ^0$ is constant with respect to the position index and assuming that $\BO$ is constructed from a single function being independently applied along the position dimension.

\paragraph{Position-equivariance} While equivariance is often a desired property, this is not true for strict position-equivariance. Identical tokens in different positions can carry vastly different implications for the correct result, as observed in Sudoku, Maze, and ARC-AGI.

\paragraph{Extrapolation beyond training cardinalities (positions and/or symbols).}
Although trained with a fixed number of positions $I$, the architecture can operate at test time on inputs with a different number of positions $J \neq I$, and some degree of generalization might be expected \citep{kazemnejad2023impact,anil2022exploring}. By contrast, it does not extrapolate to unseen symbols: problems with a larger symbol/color set $|C'|>|C|$ with $C \subset C'$ cannot be handled, as vanilla RRMs encode symbols by mapping them via  $E^{\mathcal C}$ to a high-dimensional feature space. Extrapolation to unseen symbols may  specifically be relevant in few-shot settings if several new symbols appear at test time and there are different relationships between new and old symbols.

\subsection{Symbol-equivariance: SE-RRM}

We suggest utilizing symbol equivariance in machine learning models for structured problem-solving tasks and thus extend the vanilla RRM architecture with symbol equivariance, which we denote as SE-RRMs. The core idea of how SE-RRMs are constructed is visualized in \cref{fig:main}: SE-RRM introduces a third dimension to link positions and symbols. Analogously to vanilla RRMs, SE-RRMs are based on the structurally same fixed-point operation, however with tensors of different shapes. For notational clarity and to better distinguish between the two methods, we denote the SE-RRM fixed-point operation by $\mathcal G$:
\begin{align}
\BZ^{t+1} = \mathcal G\Big(\mathbf{1}_{\text{emb}}\, E^{\mathcal G}(\BX) + \mathbf{1}_{\text{prev}}\, \BZ^{t-r}, \, \BZ^t\Big)
\end{align}
\paragraph{Task Embedding.} For SE-RRMs the symbol embedding function $E^{\mathcal C}$ is changed to realize a mapping $E^{\mathcal C}: C^I \mapsto \reals^{D \times I \times K }$, with $D$ being the number of features, $I$ being the number of positions, and $K\equiv|C|$ being the total number of symbols. Note, that in contrast to vanilla RRMs, in principle, we use the same embedding $\mathbf{d} \in \reals^D$ for all of the different symbols across all $I$ positions. An exception are special symbols like a mask or unknown token, for which we use an own embedding ($\mathbf{s}_1,\ldots,\mathbf{s}_n$). If a symbol is not present at a certain position, we use a zero embedding ($\mathbf{0}$). Formally this can be considered a function $h$ realizing:
\begin{align*}
&h : C^I \times C \times [I] \rightarrow \reals^D\\
&h(\BX, c, i) =
\begin{cases}
\mathbf{d} \in \reals^D, & \text{if } x_i=c\,\,\&\,\,c \in C_\text{usual} \\
\mathbf{s}_1 \in \reals^D, & \text{if } x_i=c\,\,\&\,\,c=s_1 \\
\vdots  & \\
\mathbf{s}_n \in \reals^D,    & \text{if } x_i=c\,\,\&\,\,c=s_n \\
\mathbf{0} \in \reals^D,    & \text{if } x_i\neq c \\
\end{cases}
\end{align*}
with $\BX = (x_1, \ldots, x_I)$, \( C = C_{\text{usual}} \cup \{ s_1, \ldots, s_n \} \), where \( s_1, \ldots, s_n \) denote special symbols and \( C_{\text{usual}} \) contains all remaining (usual) symbols.
  $h$ is uniformly applied across all $I$ positions and across all $K$ symbols. Thereby $\mathbf{d}, \mathbf{s}_1, \ldots, \mathbf{s}_n$ are constant vectors.

The extension of $E^{\mathcal H}$ to $E^{\mathcal G}$ for encoding task-type information is different for SE-RRMs compared to vanilla RRMs. While we make use of an additional position to encode task types in vanilla RRMs, we learn a task-type embedding $\mathbb{N} \rightarrow \mathbb{R}^{D \times 1 \times K}$ for SE-RRMs. This is broadcasted along the position dimension, giving an effective task-type embedding $E^{\mathcal T} : C^I \times \mathbb{N} \rightarrow \mathbb{R}^{D \times I \times K}$, which is added to the other embeddings, i.e., $E^{\mathcal G} := E^{\mathcal C} + E^{\mathcal P}+E^{\mathcal T}$. We initialize the task-type embeddings to the same vector for all $K$ symbols. The identical initialization ensures symbol equivariance at the beginning of learning, but allows to deviate for specific task types, where symbol equivariance is not appropriate at the level of single tasks of these specific task types (symbol equivariance would only hold for the whole task type in these cases). 

\paragraph{Adaptation of the output mapping.} In SE-RRM we generate predictions by linearly mapping 
the feature dimension to a single logit, thus 
reducing it to a shape of $(1,I,K)$ or $(I,K)$, from which we can either extract the prediction symbol for a position or compute a categorical loss.

\paragraph{Neural Architecture of SE-RRM blocks.} The  initial layer $\BH_0$ is analogous to RRM blocks with SE-RRM embeddings and SE-RRM recurrent states being used. In contrast to a single transformer-style layer per block in vanilla RRM, SE-RRM comprises two Transformer-style self-attention layers, where one is applied along the position dimension, while the other is applied along the symbol dimension
\citep{rao2021msa,tolstikhin2021mlp,ho2019axial,huang2019ccnet}. The update scheme $\mathcal{G}$ is then as follows:
\begin{align} 
    \BH'_{l} &= \text{Norm}\!\left( \BH_{l} + T^{D,I}(\BH_{l}) \right), &  \underbar{D} \times  \underbar{I} \times K \label{eq:serrmb1}\\
    \BH''_{l} &= \text{Norm}\!\left(\BH'_{l} + T^{D,K}(\BH'_{l})\right), &  \underbar{D} \times  I \times \underbar{K} \label{eq:serrmb2}\\
    \BH_{l+1} &= \text{Norm}\!\left(\BH''_{l} + m^{D}(\BH''_{l})\right), & \underbar{D} \times I \times K\label{eq:serrmb3}
\end{align}
for $L$ additional steps, producing the block output $\BZ^{t+1} = \BH_L$. The expressions written to the right of each equation (e.g., $\underbar{D} \times  \underbar{I} \times K$) indicate the dimensions of the 
representations, and along which dimension self-attention is applied. $T^{D,I}$ indicates that the self-attention layer is applied along the position dimension with the token dimension being the feature dimension. $T^{D,K}$ is applied along the symbol dimension with the token dimension being the feature dimension. As for vanilla RRMs, RMSNorm \citep{zhang2019root} is used as a normalization layer, which is applied uniformly across all symbols and positions.

\begin{table*}[t]
\centering
\caption{Sudoku performance of different methods across grid sizes. We report \emph{fully solved rate (FSR)}, i.e., the percentage of puzzles solved exactly, and \emph{grid-point accuracy GPA}, i.e., the percentage of correctly predicted unfilled Sudoku cells, for each mode with 95\% Wilson score confidence intervals.
All models except GPT-OSS \citep{agarwal2025gpt} are trained on $9\times9$ Sudoku grids.
The best RRM is marked bold. SE-RRM generalizes to larger
problem sizes, while the other RRMs are unable to accommodate unseen symbols.}

\label{tab:sudoku_results}

\begin{threeparttable}
\setlength{\tabcolsep}{6pt}
\renewcommand{\arraystretch}{1.15}
\resizebox{\linewidth}{!}{%
\begin{tabular}{lcccccccc}
\toprule
& \multicolumn{2}{c}{4$\times$4 Mini Sudoku} 
& \multicolumn{2}{c}{9$\times$9 Sudoku} 
& \multicolumn{2}{c}{16$\times$16 Maxi Sudoku} 
& \multicolumn{2}{c}{25$\times$25 Ultra Sudoku} \\
\cmidrule(lr){2-3}\cmidrule(lr){4-5}\cmidrule(lr){6-7}\cmidrule(lr){8-9}
Model 
& FSR & GPA
& FSR & GPA
& FSR & GPA
& FSR & GPA \\
\midrule
HRM 
& $0^{(0-1.32)}$ & $29.19^{(27.45-30.95)}$
& $63.53^{(63.38-63.67)}$ & $86.11^{(86.10-86.12)}$
& --\tnote{a} & --\tnote{a} & --\tnote{a} & --\tnote{a} \\

TRM 
& $0^{(0-1.32)}$ & $45.88^{(43.96-47.79)}$
& $71.94^{(71.80-72.08)}$ & $89.80^{(89.79-89.81)}$
& --\tnote{a} & --\tnote{a} & --\tnote{a} & --\tnote{a} \\

\textbf{SE-RRM (ours)}
& ${\bf 95.46}^{(92.01-97.08)}$ & ${\bf 99.15}^{(98.67-99.41)}$
& ${\bf 93.73}^{(93.66-93.81)}$ & ${\bf 97.58}^{(97.54-97.63)}$
& $0^{(0-1.75)}$ & ${\bf 51.95}^{(51.43-52.47)}$
& $0^{(0-8.38)}$ & ${\bf 31.49}^{(30.22-32.75)}$ \\ \midrule

GPT-OSS-20B\tnote{b} 
& $100^{(98.68-100)}$ & $100^{(99.85-100)}$
& $21.67^{(13.12-33.62)}$ & --\tnote{c}
& --\tnote{d} & --\tnote{d} & --\tnote{d} & --\tnote{d} \\
\bottomrule
\end{tabular}
}

\begin{tablenotes}
\footnotesize
\item[a] Cannot extrapolate to this size.
\item[b] LLM-based systems are evaluated without task-specific program synthesis or symbolic search.\\
\item[c] The model sometimes refuses to produce any output, thus GPA is therefore not computed.  \item[d] Model refuses to provide solutions.
\end{tablenotes}
\end{threeparttable}
\end{table*}

As for vanilla RRMs, when omitting positional and task-type encodings in the task embeddings, position equivariance can be shown.

\begin{proposition} \label{serrmposeq}
    An SE-RRM block $\mathcal G$ without positional embeddings, as defined in~\cref{eq:rrmb0,eq:serrmb1,eq:serrmb2,eq:serrmb3} is equivariant under permutations $\pi: [I] \mapsto [I]$ of the input positions $\in [I]:=\{1,\ldots,I\}$, i.e.,
\begin{align*}
\Pi_2^\pi\Big(\mathcal G\Big(\mathbf{1}_{\text{emb}}\, E^{\mathcal G}(\BX) + \mathbf{1}_{\text{prev}}\, \BZ^{t-r}, \, \BZ^t\Big)\Big) = \nonumber \\
=\mathcal G\Big(\mathbf{1}_{\text{emb}}\, E^{\mathcal G}(\Pi_2^\pi(\BX)) + \mathbf{1}_{\text{prev}}\, \Pi_2^\pi(\BZ^{t-r}), \, \Pi_2^\pi(\BZ^t)\Big),
\end{align*}
where $\Pi_a^\pi$ is defined as for \cref{RRMposeq}.

\end{proposition}

The proof is analogous to \cref{RRMposeq}, with the only differences being the addition of an extra dimension and the application of a self-attention operation along the symbol dimension in \cref{eq:serrmb2}. \cref{eq:serrmb2} is uniformly applied to all positions and therefore does not break position equivariance. Since all other operations are further uniform across the additional symbol dimension, position equivariance is preserved.

Furthermore, the SE-RRM block exhibits the beneficial property of equivariance under the permutations of symbols.
\begin{proposition}
    An SE-RRM block $\mathcal G$ without task-type embeddings, as defined in~\cref{eq:rrmb0,eq:serrmb1,eq:serrmb2,eq:serrmb3} is equivariant under
    permutations $\rho: [K] \mapsto [K]$ of the input symbols $\in [K]:=\{1,\ldots,K\}$, i.e.,
\begin{align*}
\Pi_3^\rho\Big(\mathcal G\Big(\mathbf{1}_{\text{emb}}\, E^{\mathcal G}(\BX) + \mathbf{1}_{\text{prev}}\, \BZ^{t-r}, \, \BZ^t\Big)\Big) = \nonumber \\
=\mathcal G\Big(\mathbf{1}_{\text{emb}}\, E^{\mathcal G}(\Pi_3^\rho(\BX)) + \mathbf{1}_{\text{prev}}\, \Pi_3^\rho(\BZ^{t-r}), \, \Pi_3^\rho(\BZ^t)\Big),
\end{align*}
where $\Pi_a^\pi$ is defined as for \cref{RRMposeq}.
\end{proposition}

The proof is analogous to \cref{serrmposeq} with adaptations according to the application to another dimension. The inclusion of a symbol dimension with an appropriate embedding allows this equivariance.  

\paragraph{Computational Complexity.}
An efficient implementation of Attention \citep{flashattn} has a computational complexity of $O(I^2)$ and a memory complexity of $O(I)$. For the attention mechanism in SE-RRM this changes to $O(I^2K + K^2I)$ and $O(IK)$ respectively. For the problems considered $I \gg K$, therefore SE-RRM increases computational and memory complexity linearly by the factor $K$.  Conversely, in domains where $K \gg I$, the computational complexity is dominated by 
$O(K^2I)$, which may render SE-RRM impractical.

\section{Experiments and Results}
We evaluate Hierarchical Reasoning Models (HRMs), Tiny Recursive Models (TRMs), 
and our proposed Symbol-Equivariant Recurrent Reasoning Models (SE-RRMs) 
on three classes of structured problem-solving tasks: Sudoku, ARC-AGI, and Maze-solving. 
All recurrent reasoning models share the same overall training protocol established in HRM \citep{wang2025hierarchical}: 
models are trained end-to-end using supervised learning with deep supervision 
at each recurrent step, and inference is performed by repeatedly 
applying the same recurrent block for a fixed number of iterations. 
Large language model baselines are evaluated without 
task-specific program synthesis, symbolic search, or 
human-designed orchestration, reflecting their performance as plain neural solvers.
HRM and TRM employ a Q-learning approach that serves as a stopping policy for deep supervision, allowing the model to apply fewer than the maximum number of deep supervision steps during training and thereby reducing computation time. In contrast, SE-RRM uses a simpler stochastic mechanism: at each deep supervision step, training-time supervision is terminated with probability pp
p (set as a hyperparameter), unless the maximum number of steps has already been reached. During inference, all models — HRM, TRM, and SE-RRM — execute the full maximum number of deep supervision steps, which is set to 16 for all experiments in this section.

In addition, we apply RoPE2d \citep{heo2024rotary} instead of RoPE to encode the spatial positions $I$.
A detailed comparison of all hyperparameters together with various ablations can be found in the appendix. Throughout the experiments we use an SE-RRM with only 2 million parameters, which is significantly smaller than the compared methods.

\subsection{Sudoku}
\paragraph{Data.} We use a subset of 1,000 samples of the $9\times9$ Sudoku dataset provided by \citet{wang2025hierarchical} (HRM), compute 1,000 augmentations per training sample like HRM and TRM, and use the full test set of 422,786 samples for validation.
In addition, we test zero-shot generalization to smaller and larger Sudoku puzzles.
To this end, we consider $4\times 4$, $16\times 16$, and $25\times 25$ puzzles.
For $4\times 4$ puzzles, the test set comprises all 288 unique combinations\footnote{\url{https://github.com/Black-Phoenix/4x4-Sudoku-Dataset}}, while we created 216 puzzles with unique solutions of varying difficulty with \texttt{py-sudoku} for $16\times16$ Sudoku.
Finally, we extracted $42$ samples of size $25\times 25$ from the curated puzzles in \texttt{Sudoku-SMT-Solvers}\footnote{\url{https://github.com/liamjdavis/Sudoku-SMT-Solvers}. Note that the solver \texttt{Z3Solver} solved only 42 out of 100 puzzles.}.

\paragraph{Approach.} We evaluate all models on Sudoku puzzles of varying grid sizes that range from small $4\times4$ instances to large $25\times25$ puzzles. 
All RRMs (HRM, TRM, and SE-RRM) are trained exclusively on standard $9\times9$ Sudokus and tested without fine-tuning on other grid sizes to assess generalization and extrapolation. 
Performance is measured using two metrics: \emph{fully solved rate (FSR)}, defined as the fraction of puzzles solved exactly, and \emph{grid-point accuracy (GPA)}, defined as the fraction of correctly predicted, \emph{initially unfilled} cells. We report 95\% confidence intervals using the Wilson score interval. 

We use the hyperparameters of the respective papers, thus performing 16 deep supervision steps on inference for HRM, TRM and also SE-RRM. An ablation of the impact of the number of steps on the performance can be found in the appendix.

\paragraph{Results.} The quantitative evaluation is summarized in Table~\ref{tab:sudoku_results}.
Here, 
At the training regime ($9\times9$), both HRM and TRM achieve strong performance and outperform GPT-OSS-20B \citep{agarwal2025gpt}.
GPT-OSS-20B -- having no access to external tools -- fails to reliably solve even standard $9\times9$ puzzles despite having been exposed to Sudoku-like problems during pretraining.
Our proposed SE-RRM model outperforms all base-lines by a large margin ($>11\%$ for FSR and $>7\%$ for GPA) proving that reasoning using symbol equivariance is more effective.
Note that TRM~\citep{jolicoeur2025less} reports an accuracy of 87.4 \% on Sudoku with an MLP-mixer~\cite{tolstikhin2021mlp}, however this version supports only a fixed number of positions and is thus unable to extrapolate to different grids.

Next, we evaluate how the models generalize to simpler $4\times 4$ puzzles.
HRM and TRM do not extrapolate to $4\times 4$ mini Sudoku as the FSR rate drops to zero, 
showing that both models did not learn the intrinsic Sudoku rules.
In contrast, GPT-OSS-20B can perfectly solve these simple logic puzzles.
Remarkably, SE-RRM is able to extrapolate the Sudoku rules to $4\times 4$ grids and achieves $95.46\%$ FSR and gets almost all unfilled cells right (GPA $99.15\%$).

Finally, we test generalization to larger grids (16×16 and 25×25), with $I$ (number of positions) growing quadratically and $K$ (number of symbols) growing linearly.
HRM and TRM cannot be applied to larger Sudoku grids without any training, as additional symbols ("10" up to "25") would require novel embeddings.
Despite numerous attempts, GPT-OSS-20B did not provide any meaningful outputs.
This result, along with the performance on $9\times 9$ grids, corroborates prior observations that LLMs struggle with tightly constrained symbolic reasoning tasks.
Although the trained SE-RRM model does not solve any of the larger Sudokus, the GPA of 51.95\% for $16\times 16$ grids and 31.49\% for $25\times25$ grids is substantially higher than random choices and demonstrates that our trained model is able to generalize beyond the training data distribution.

\paragraph{Test time scaling.} RRMs offer the possibility to adjust the number of steps during inference. For a NP-hard problem like Sudoku, this can improve the performance of the model substantially. Table~\ref{tab:test_time_scaling} shows that SE-RRM does not only outperform the other RRMs on a different number of deep supervision steps, but also delivers overall the best performance.

\begin{table}[h]
\centering
\caption{Test-time scaling on $9\times9$ Sudoku. Fully solved rate (FSR in \%) as a function of the number of deep supervision steps at inference.}
\label{tab:test_time_scaling}
\setlength{\tabcolsep}{5pt}
\renewcommand{\arraystretch}{1.15}
\resizebox{\linewidth}{!}{
\begin{tabular}{lcccccccc}
\toprule
& \multicolumn{8}{c}{Number of inference steps} \\
\cmidrule(lr){2-9}
Model & 1 & 2 & 4 & 8 & 16 & 32 & 64 & 128 \\
\midrule
HRM   & $2.01$  & $20.47$ & $49.37$ & $59.45$ & $63.53$ & $65.67$ & $67.09$ & $68.15$ \\
TRM   & $2.51$  & $31.02$ & $55.49$ & $66.46$ & $71.95$ & $75.36$ & $77.86$ & $79.78$ \\
\textbf{SE-RRM (ours)} & $\mathbf{16.05}$ & $\mathbf{62.06}$ & $\mathbf{77.31}$ & $\mathbf{87.38}$ & $\mathbf{93.73}$ & $\mathbf{96.82}$ & $\mathbf{98.22}$ & $\mathbf{98.84}$ \\
\bottomrule
\end{tabular}
}
\end{table}

\begin{table}[h]
\centering
\caption{pass@2 on ARC-AGI benchmarks and fully solved rate (FSR) on Maze with 95\% Wilson score confidence intervals.
Reported values for HRM and TRM are taken from the respective papers.}
\label{tab:arc_agi_maze_results}
\setlength{\tabcolsep}{6pt}
\renewcommand{\arraystretch}{1.15}
\resizebox{\linewidth}{!}{%
\begin{tabular}{lccc}
\toprule
& ARC-AGI-1 & ARC-AGI-2 & Maze \\
\cmidrule(lr){2-3}\cmidrule(lr){4-4}
Model & \multicolumn{2}{c}{pass@2} & FSR \\
\midrule
HRM & $40.3^{(35.56-45.13)}$ & $5.0^{(2.31-10.48)}$ & $74.5^{(71.66-77.15)}$ \\
TRM & {44.6}$^{(39.71-49.40)}$ & \textbf{7.8}$^{(4.0-13.64)}$ & $85.3^{(82.92-87.41)}$ \\
SE-RRM (ours) & $\textbf{45.3}^{(40.44-50.15)}$  & {7.1}$^{(3.42-12.61)}$ & \textbf{88.8}$^{(86.64-90.65)}$ \\
\bottomrule
\end{tabular}
}
\end{table}

\subsection{ARC-AGI}
\paragraph{Data.} Throughout this section, we use the official training and evaluation sets of ARC-AGI-1 \cite{chollet2019measure} and ARC-AGI-2 \cite{chollet2025arc}, consisting of geometric puzzles.
Each puzzle uses at most 10 colors distributed across a grid of at most $30\times30$ positions and 2-11 few-shot examples demonstrate the underlying task.
Like HRM and TRM, we include 160 puzzles of the ConceptARC-dataset \citep{conceptarc} and all few-shot examples of the evaluation puzzles for training.
Therefore, the training set comprises 960 and 1,280 puzzles for ARC-AGI-1 and ARC-AGI-2, respectively.
The resulting models are evaluated on the official 400 test puzzles for ARC-AGI-1 and 120 test puzzles of the ARC-AGI-2 challenge.

\paragraph{Approach.} We compare the officially reported performance values of HRM, TRM, and SE-RRM. In contrast to the extensive color augmentations applied by HRM and TRM, we only use 8 dihedral augmentations for each puzzle due to symbol equivariance of SE-RRM. Moreover, several ablations, where we incrementally apply the hyperparameters of SE-RRM to TRM, are presented in the appendix.

\paragraph{Results.} SE-RRM outperforms HRM on both ARC-AGI-1 and ARC-AGI-2 and delivers comparable results to TRM. In Table~\ref{tab:arc_agi_maze_results}, we report results from the original publications of HRM and TRM, compared with SE-RRM. 

\subsection{Maze}
\paragraph{Data.} Finally, we evaluate the models on grid-based maze-solving tasks, which require multi-step planning, global constraint satisfaction, 
and consistent state tracking.
All models are trained on the Maze-hard dataset defined by \citet{wang2025hierarchical}, consisting of 1,000 training and 1,000 test mazes.
Each maze is defined on a $30\times 30$ grid and has a minimal solution path length of 110.
Note that we do not perform any augmentations during training of SE-RRM.

\paragraph{Approach.}
In contrast to the previous puzzles, symbol equivariance is not reasonable for this task, as walls are not equivalent to start and end points.
Thus, we allow different embeddings for each symbol, breaking the symbol equivariance. The resulting model is neither equivariant towards permutations of symbols nor to permutations of positions.

\paragraph{Results.}
Consistent with the results on Sudoku and ARC-AGI, 
SE-RRM achieves competitive results on Maze (Table~\ref{tab:arc_agi_maze_results}).
SE-RRM with broken symbol-equivariance performs slightly better than TRM and much better than HRM. This could be due to the higher-dimensional input embedding or the additional transformer layer in an SE-RRM block. In any case, it shows that SE-RRM can be successfully applied to problems that do not require symbol equivariance.

\section{Conclusion and Limitations}
Overall, across all three domains, we show that symbol-equivariant recurrent reasoning models offer a robust and scalable alternative to both symbolic solvers and LLM-based approaches for structured problem-solving tasks. We show that SE-RRMs require substantially fewer augmentations. Specifically, we trained on ARC-AGI with 8 instead of 1000 augmentations per sample. On Sudoku, where no compared model applies extensive permutations of symbols as data augmentation, SE-RRM delivers superior results. Moreover, our architecture enables recurrent reasoning models to accommodate novel symbols, which allows extrapolation to different task formulations.

\paragraph{Limitations.} Throughout the paper, we only 
used SE-RRMs with 2M parameters. This is substantially 
fewer than TRM (7M) or HRM (27M). The reduced parameter 
count can partially offset the increased computational 
and memory complexity during training and inference.





\section*{Acknowledgements}
We thank Sebastian Lehner for helpful discussions.
The ELLIS Unit Linz, the LIT AI Lab, the Institute for Machine 
Learning, are supported by the Federal State Upper Austria. 
We thank the projects FWF Bilateral Artificial Intelligence (10.55776/COE12), FWF AIRI FG 9-N (10.55776/FG9), 
AI4GreenHeatingGrids (FFG- 899943), Stars4Waters (HORIZON-CL6-2021-
CLIMATE-01-01). We also thank NXAI GmbH, 
Audi AG, Silicon Austria Labs 
(SAL), Merck Healthcare KGaA, GLS (Univ. Waterloo), T\"{U}V 
Holding GmbH, Software Competence Center Hagenberg GmbH, dSPACE 
GmbH, TRUMPF SE + Co. KG.


\section*{Impact Statement}
This paper presents work whose goal is to advance the field of Machine
Learning. There are many potential societal consequences of our work, none
which we feel must be specifically highlighted here.

\bibliography{bib}
\bibliographystyle{icml2026}


\appendix
\setcounter{figure}{0}
\setcounter{table}{0}
\renewcommand{\thefigure}{A\arabic{figure}}
\renewcommand{\thetable}{A\arabic{table}}
\onecolumn

\section{Notation}

\begin{table}[H]
\centering
\caption{Notation used in the description of Recurrent Reasoning Models (RRMs), Hierarchical Reasoning Models (HRMs), Tiny Recursive Models (TRMs), and Symbol-Equivariant RRMs (SE-RRMs).\label{tab:notation}}
\resizebox{0.80 \textwidth}{!}{%
\begin{tabular}{l c l}
\toprule
\textbf{Definition} & \textbf{Symbol} & \textbf{Type} \\
\midrule

\multicolumn{3}{l}{\textbf{Scalars and indices}} \\
feature dimension & $D$ & $\mathbb{N}$ \\
number of positions (grid points) & $I$ & $\mathbb{N}$ \\
number of symbols / colors & $K$ & $\mathbb{N}$ \\
recurrent iteration index & $t$ & $\mathbb{N}$ \\
number of layers per block & $L$ & $\mathbb{N}$ \\
final iteration step & $T$ & $\mathbb{N}$ \\
number of additional task-type positions (RRM) & $P$ & $\mathbb{N}$ \\

\midrule
\multicolumn{3}{l}{\textbf{Sets and spaces}} \\
symbol / color alphabet & $C$ & finite set, $|C| = K$ \\
input and solution space & $C^I$ & tuples of symbols \\

\midrule
\multicolumn{3}{l}{\textbf{Tasks and solutions}} \\
input task & $\BX = (x_1,\ldots,x_I)$ & $C^I$ \\
solution task & $\BY = (y_1,\ldots,y_I)$ & $C^I$ \\
symbol at position $i$ & $x_i,\, y_i$ & $C$ \\
task-type identifier & $p$ & $\mathbb{N}$ \\

\midrule
\multicolumn{3}{l}{\textbf{Embeddings and representations}} \\
symbol-level embedding (SE-RRM) & $E^{\mathcal C}$ & $C^I \mapsto \mathbb{R}^{D \times I \times K}$ \\
positional embedding (SE-RRM) & $E^{\mathcal P}$ & $C^I \to \mathbb{R}^{D \times I \times K}$ \\
task-type embedding (SE-RRM) & $E^{\mathcal T}$ & $C^I \times \mathbb{N} \rightarrow \mathbb{R}^{D \times 1 \times K}$\\
combined task embedding (SE-RRM) & $E^{\mathcal G}$ & $E^{\mathcal C} + E^{\mathcal P} + E^{\mathcal T}$ \\
combined task embedding (RRM) & $E^{\mathcal H}$ & $C^I \mapsto \mathbb{R}^{D \times I}$ or $C^I \times \mathbb{N} \rightarrow \mathbb{R}^{D \times (I+P)}$ \\
SE-RRM symbol embedding function & $h$ & $C^I \times C \times [I] \rightarrow \mathbb{R}^D$ \\
shared usual-symbol embedding vector & $\mathbf{d}$ & $\mathbb{R}^D$ \\
special-symbol embedding vectors & $\mathbf{s}_1, \ldots, \mathbf{s}_n$ & $\mathbb{R}^D$ \\

HRM/TRM task embedding matrix & $\BE=E^{\mathcal H}(\BX)$ & $\mathbb{R}^{D \times I}$ \\
SE-RRM task embedding tensor & $\BE'=E^{\mathcal G}(\BX)$ & $\mathbb{R}^{D \times I \times K}$ \\
token embedding at position $i$ & $\Be_i$ & $\mathbb{R}^{D}$ \\
intermediate block states & $\BH_l, \BH'_l,\BH_l''$ &   $\mathbb{R}^{D \times I}$ or $\mathbb{R}^{D \times I \times K}$\\
recurrent state at step $t$ & $\BZ^t$ & $\mathbb{R}^{D \times I}$ or $\mathbb{R}^{D \times I \times K}$ \\

\midrule
\multicolumn{3}{l}{\textbf{Models and blocks}} \\
RRM block (HRM/TRM) & $\mathcal H$ & $\mathbb{R}^{D \times I} \times \mathbb{R}^{D \times I} \mapsto \mathbb{R}^{D \times I}$ \\
SE-RRM block & $\mathcal G$ & $\mathbb{R}^{D \times I \times K} \times \mathbb{R}^{D \times I \times K} \mapsto \mathbb{R}^{D \times I \times K}$ \\
embedding inclusion indicator & $\mathbf{1}_{\text{emb}}$ & $\{0, 1\}$ \\
previous-state inclusion indicator & $\mathbf{1}_{\text{prev}}$ & $\{0, 1\}$ \\

\midrule
\multicolumn{3}{l}{\textbf{Neural operators}} \\
self-attention over positions & $T^{D,I}$ & $\mathbb{R}^{D \times I} \mapsto \mathbb{R}^{D \times I}$ \\
self-attention over symbols & $T^{D,K}$ & $\mathbb{R}^{D \times K} \mapsto \mathbb{R}^{D \times K}$ \\
MLP over features & $m^{D}$ & $\mathbb{R}^{D} \mapsto \mathbb{R}^{D}$ \\
normalization (RMSNorm) & $\mathrm{Norm}$ & $\mathbb{R}^{D} \mapsto \mathbb{R}^{D}$  \\

\midrule
\multicolumn{3}{l}{\textbf{Prediction and loss}} \\
output mapping (RRM) & $\BO$ & $\mathbb{R}^{D \times I} \mapsto \mathbb{R}^{K \times I}$ \\
output projection matrix (RRM) & $\BW$ & $\mathbb{R}^{K \times D}$ \\
output projection matrix (SE-RRM) & $\BW$ & $\mathbb{R}^{1 \times D}$ \\
logits at step $t$ & $\hat{\BY}^t$ & $\mathbb{R}^{K \times I}$ (RRM) or $\mathbb{R}^{I \times K}$ (SE-RRM) \\
categorical cross-entropy loss & $L(\hat{\BY}, \BY)$ & $\mathbb{R}_{\ge 0}$ \\

\midrule
\multicolumn{3}{l}{\textbf{Symmetry operators}} \\
permutation of positions & $\pi$ & bijection on $\{1,\ldots,I\}$ \\
permutation of symbols/colors & $\rho$ & bijection on $\{1,\ldots,K\}$ \\
permutation operator on tensor & $\Pi_a^\pi$ & permutes axis $a$ with permutation $\pi$ \\

\midrule
\multicolumn{3}{l}{\textbf{Task-type components}} \\
task-type embedding position (RRM) & $\Bp$ & $\mathbb{R}^{D}$ \\
task-type embedding matrix (SE-RRM) & $\BP$ & $\mathbb{R}^{D \times K}$ \\

\end{tabular}
}
\end{table}

\section{Symbol Equivariance}

\begin{figure}[th]
    \centering
    \includegraphics[width=8cm]{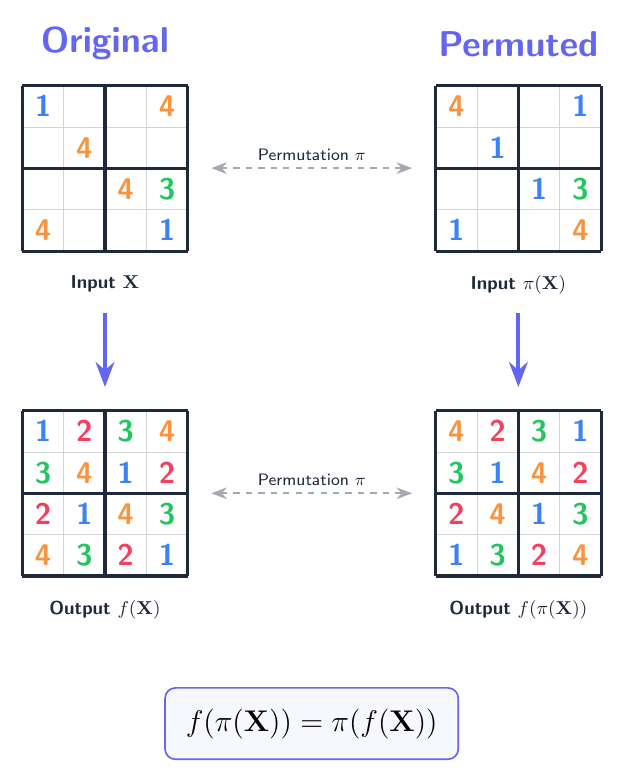}
    \caption{Illustration of Model Equivariance under Symbol Permutation. The model f is equivariant if applying a permutation $\pi$ to the input $X$ results in an equivalent permutation of the output, such that $f(\pi(X))=\pi(f(X))$. Here, the permutation $\pi$ swaps the symbols 1 and 4. In contrast to other reasoning models, SE-RRM guarantees permutation equivariance.}
    \label{fig:sudoku_permutation}
\end{figure}

\begin{figure}[th]
    \centering
    \includegraphics[width=8cm]{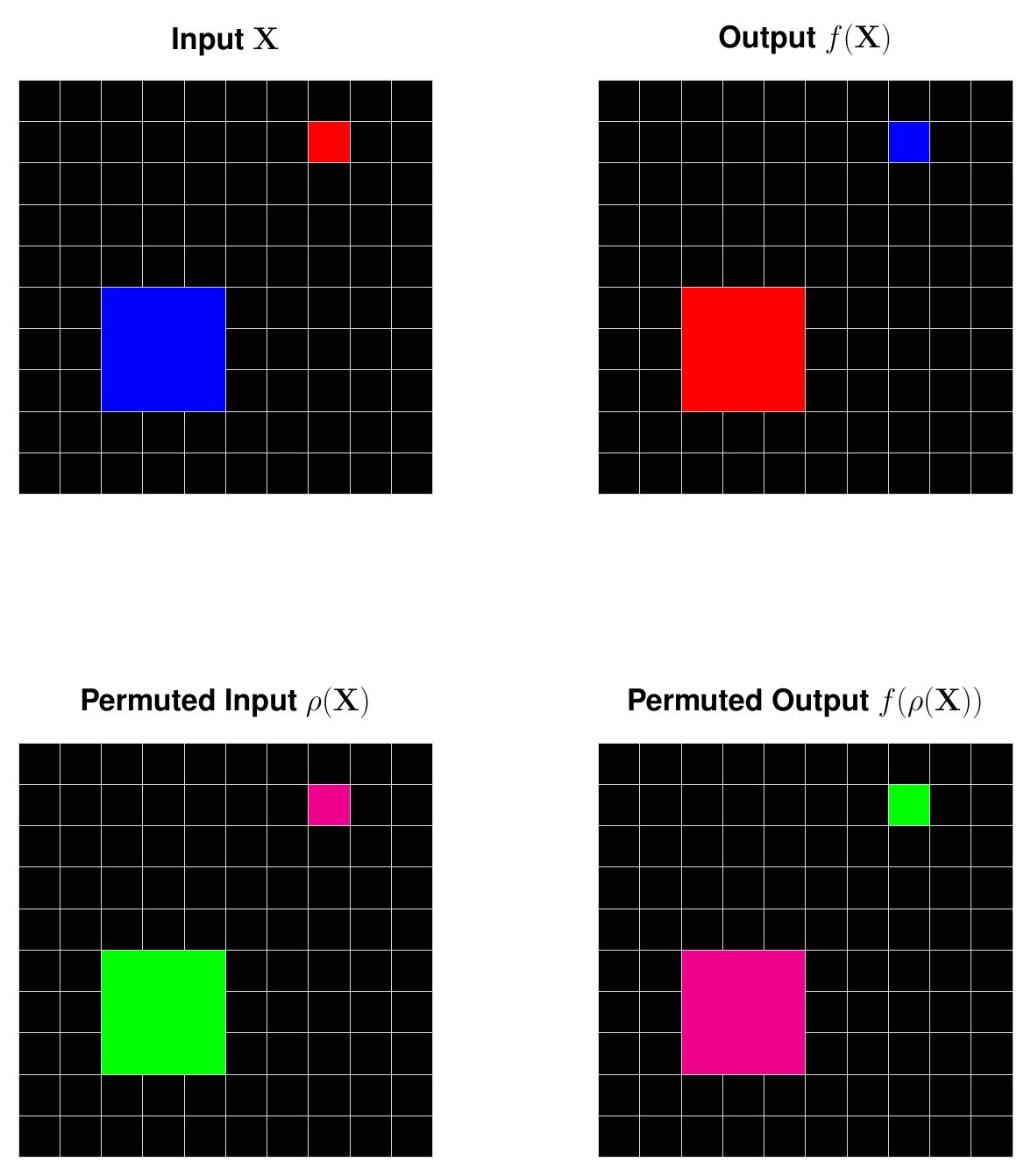}
    \caption{Illustration of an exemplary ARC-AGI task where the colors of the squares in the input have to be exchanged. Vanilla RRMs would rely on extensive colour augmentations, while SE-RRMs can immediately learn the task type rule.}
    \label{fig:arc}
\end{figure}

Based on a small $4 \times 4$ Sudoku we intuitively explain the concept of symbol equivariance in Figure~\ref{fig:sudoku_permutation}. Moreover, we show a simple exemplary ARC-AGI example in Figure~\ref{fig:arc}. 

\clearpage

\section{Architecture and hyperparameters}

\subsection{Sudoku}

\begin{table}[H]
\centering
\caption{Hyperparameter Comparison for Sudoku}
\label{tab:hyperparams_sudoku}
\begin{tabular}{@{}llll@{}}
\toprule
\textbf{Hyperparameter} & \textbf{HRM} & \textbf{TRM}  & \textbf{SE-RRM}\\ 

\midrule
\textit{Deep supervision steps} & 16 & 16  & 16 \\
\textit{early stopping of deep supervision} & q-learning & q-learning  & random (5\% chance per step) \\
\textit{Recursive Cycles} & $H=2, L=2$ & $H=3, L=6$  & $H=3, L=6$  \\
\textit{Batch Size} & 384 & 768  & 272 \\
\textit{Number of epochs} & 50000 & 50000  & 10000 \\
\textit{Learning Rate} & 0.0001 & 0.0001  & 0.0005 \\
\textit{Weight Decay} & 1 & 1 & 1 \\
\textit{LR Schedule} & Warmup + Constant-LR & Warmup + Constant-LR & Warmup + Constant-LR \\
\textit{Hidden Size} & 512 & 512 & 256 \\
\textit{Parameters} & 27M & 7M & 2M \\
\textit{Transformer Layers} & 8 & 2 & 4 (2 for each dimension) \\ 
\textit{Positional Encoding} & RoPE &RoPE & RoPE2D \\ 
\bottomrule
\end{tabular}
\end{table}

\subsection{ARC-AGI}

\begin{table}[H]
\centering
\caption{Hyperparameter Comparison for ARC-AGI. All models use the same hyperparameters for ARC-AGI-1 and ARC-AGI-2.}

\label{tab:hyperparams_arc}
\begin{tabular}{@{}llll@{}}
\toprule
\textbf{Hyperparameter} & \textbf{HRM} & \textbf{TRM}  & \textbf{SE-RRM}\\ 

\midrule
\textit{Deep supervision steps} & 16 & 16  & 16 \\
\textit{early stopping of deep supervision} & q-learning & q-learning  & random (10\% chance per step) \\
\textit{Recursive Cycles} & $H=2, L=2$ & $H=3, L=4$  & $H=3, L=4$  \\
\textit{Batch Size} & 768 & 768  & 272 \\
\textit{Number of epochs} & 100000 & 100000  & 100000 \\
\textit{Learning Rate} & 0.0001 & 0.0001  & 0.0005 \\
\textit{Weight Decay} & 0.1 & 0.1 & 0.1 \\
\textit{LR Schedule} & Warmup + Constant-LR & Warmup + Constant-LR & Warmup + Cosine-Decay \\
\textit{Learning Rate for puzzle embeddings} & 0.01 & 0.01  & 0.01 \\
\textit{Weight Decay for puzzle embeddings} & 0.1 & 0.1  & 0.3 \\
\textit{Hidden Size} & 512 & 512 & 256 \\
\textit{Parameters} & 27M & 7M & 2M \\
\textit{Transformer Layers} & 8 & 2 & 4 (2 for each dimension) \\ 
\textit{Positional Encoding} & RoPE &RoPE & RoPE2D \\ 
\bottomrule
\end{tabular}
\end{table}

\subsection{Maze}

\begin{table}[H]
\centering
\caption{Hyperparameter Comparison for Maze}
\label{tab:hyperparams_maze}
\begin{tabular}{@{}llll@{}}
\toprule
\textbf{Hyperparameter} & \textbf{HRM} & \textbf{TRM}  & \textbf{SE-RRM}\\ 

\midrule
\textit{Deep supervision steps} & 16 & 16  & 16 \\
\textit{early stopping of deep supervision} & q-learning & q-learning  & random (10\% chance per step) \\
\textit{Recursive Cycles} & $H=2, L=2$ & $H=3, L=4$  & $H=3, L=4$  \\
\textit{Batch Size} & 384 & 768  & 64 \\
\textit{Number of epochs} & 20000 & 50000  & 10000 \\
\textit{Learning Rate} & 0.0001 & 0.0001  & 0.0005 \\
\textit{Weight Decay} & 1 & 1 & 1 \\
\textit{LR Schedule} & Warmup + Constant-LR & Warmup + Constant-LR & Warmup + Constant-LR \\
\textit{Hidden Size} & 512 & 512 & 256 \\
\textit{Parameters} & 27M & 7M & 2M \\
\textit{Transformer Layers} & 8 & 2 & 4 (2 for each dimension) \\ 
\textit{Positional Encoding} & RoPE &RoPE & RoPE2D \\ 
\bottomrule
\end{tabular}
\end{table}

\section{Additional experimental results}

\subsection{Ablation on ARC-AGI-1}

In Table~\ref{tab:arc_agi_morph}, we gradually transform TRM into SE-RRM and show the implications on the ARC-AGI-1 benchmark. First, we replace Q-learning with the random halting mechanism of deep supervision (TRM + random halting). Then, we change RoPE to RoPE2d. Interestingly, both of these changes have a clear negative effect on TRM's performance. When we additionally match all remaining hyperparameters, this effect is even more pronounced. However, this is less surprising, as SE-RRM also has significantly fewer parameters. After further adding the 2-dimensional embedding and adopting our architecture, we arrive at SE-RRM.

\begin{table}[h]
\centering
\caption{pass@2 on ARC-AGI-1 benchmark with 95\% Wilson score confidence intervals. The hyperaparameters and architecture of TRM is changed step by step into SE-RRM.}
\label{tab:arc_agi_morph}
\setlength{\tabcolsep}{6pt}
\renewcommand{\arraystretch}{1.15}
\begin{tabular}{lc}
\toprule
Model & ARC-AGI-1 (\%)\\
\midrule
TRM & {44.6}$^{(39.71-49.40)}$ \\
TRM + random halting  & {34.8}$^{(30.25-39.54)}$ \\
TRM + RoPE2d  & {
38.4
}$^{(33.62-43.10)}$ \\
TRM + random halting + RoPE2d &{33.1}$^{(28.57-37.75)}$ \\
TRM + all hyperparameters matched &{32.3}$^{(27.86-36.98)}$ \\
TRM + all hyperparameters matched + equivariance = SE-RRM (ours) &$\textbf{45.3}^{(40.44-50.15)}$  \\
\bottomrule
\end{tabular}
\end{table}

\subsection{Learning dynamics}

\begin{figure}[h]
    \centering
    \includegraphics[width = \linewidth]{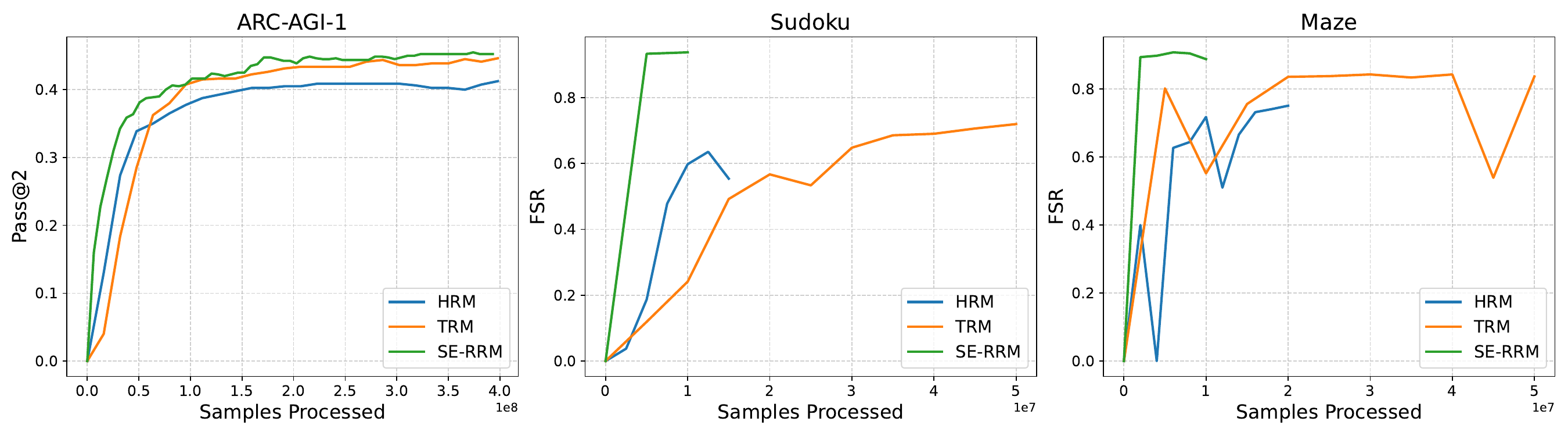}
    \caption{Learning curves for HRM, TRM, and SE-RRM. Validation metrics HRM, TRM and SE-RRM on Sudoku, ARC-AGI-1 and Maze are shown on the y-axis. The number of samples 
    processes during training is shown on the x-axis. SE-RRM is more data efficient since it reaches higher metrics with lower number of samples processed. }
    \label{fig:learning_curves}
\end{figure}

In Figure~\ref{fig:learning_curves} we compare the development of the Pass@2 accuracy on ARC-AGI1 and FSR (Fully Solved Rate) on Sudoku and Maze of HRM, SE-RRM and TRM during training on different datasets.

\subsection{Ablation on Sudoku and Maze}

We further evaluate the different choices of hyperparameters on Sudoku and Maze. More specifically, we train TRM both with random halting and q-learning, evaluate both models with RoPE and RoPE2D, and try hyperparameters of TRM on SE-RRM and the other way round. Note that we do not try q-learning with SE-RRM, as the encoding has a different shape which prevents a straight-forward implementation. Results are shown in Table~\ref{tab:ablations}. While SE-RRM seems to work quite well under different configurations, TRM sometimes does not work at all. We suspect that the hyperparameters of TRM are optimized for q-learning and don't transfer well to different settings.

In Table~\ref{tab:sudoku_maze} we ablate the effect of enforced equivariance via shared embeddings. SE-RRM performs best on Sudoku with enforced equivariance and slightly worse without. Note that equivariance can still be learned during training, even if the initial embeddings for every digit are different. In Maze, the symbols have a clearly different semantics. Therefore, enforcing equivariance leads to inferior performance.

\begin{table}[h]
\centering
\caption{Ablations of TRM and SE-RRM on Sudoku and Maze.}
\label{tab:ablations}
\begin{tabular}{lcccccc}
\toprule
Model & Halting & PosEnc & Hyperparameters & Sudoku (\%) & Maze (\%)\\
\midrule
TRM & q-learning & RoPE & TRM & $71.9^{(71.80-72.08)}$ &$85.3^{(82.92-87.41)}$ \\
TRM & random & RoPE & TRM & $79.4^{(79.31-79.55)}$ & $84.9^{(82.55-86.99)}$ \\
TRM & q-learning & RoPE2D & TRM & $76.5^{(76.34-76.59)}$ & $85.4^{(83.08-87.45)}$\\ 
TRM & random & RoPE2D & TRM & $0^{(0.00-0.00)}$ & $59.3^{(56.23-62.30)}$\\ 
TRM & random & RoPE2D & SE-RRM &  $0^{(0.00-0.00)}$ & $65.0^{(61.99-67.89)}$\\
SE-RRM & random & RoPE2D & SE-RRM & {93.7}$^{(93.66-93.81)}$ & \textbf{88.8}$^{(86.64-90.65)}$ \\
SE-RRM & random & RoPE & SE-RRM & \textbf{95.4}$^{(95.32-95.45)}$ & $78.3^{(75.64-80.74)}$ \\
SE-RRM & random & RoPE & TRM & $68.4^{(68.30-68.58)}$ &$80.1^{(77.51-82.46)}$\\ 
\bottomrule
\end{tabular}
\end{table}

\begin{table}[h]
\centering
\caption{Results of SE-RRM on Sudoku and Maze including ablations. The second column indicates whether symbol equivariance is enforced through shared embeddings.}
\label{tab:sudoku_maze}
\begin{tabular}{lccc}
\toprule
& Symbol & Sudoku & Maze \\
Model &  Equivariance & FSR &  FSR 
\\
\midrule
TRM & False & $71.9^{(71.80-72.08)}$ &  $85.3^{(82.92-87.41)}$ \\
HRM & False & $63.5^{(63.38-63.67)}$ & $74.5^{(71.66-77.15)}$ \\
SE-RRM & True &  \textbf{93.7}$^{(93.66-93.81)}$ &$0^{(0.00-0.38)}$\\
SE-RRM & False & {89.7}$^{(89.61-89.79)}$  & \textbf{88.8}$^{(86.64-90.65)}$ \\

\bottomrule
\end{tabular}
\end{table}

\end{document}